%% file: rearrangement.tex
\documentclass[conference]{IEEEtran}
\usepackage{times}

\usepackage[numbers]{natbib}
\usepackage{multicol}
\usepackage[bookmarks=true]{hyperref}

\usepackage{amsmath,amsfonts,amssymb,mathrsfs}
\usepackage{verbatim}
\let\chapter\section
\usepackage[vlined,ruled,linesnumbered]{algorithm2e}
\usepackage{wrapfig}
\usepackage{graphicx} 
\usepackage{xspace}
\usepackage{subfigure}

\begin{document}

\title{Similar Part Rearrangement in Clutter\\
With Pebble Graphs}

\author{\authorblockN{Athanasios Krontiris, Rahul Shome, Andrew Dobson,
 Andrew Kimmel, Isaac Yochelson, Kostas E. Bekris}
\authorblockA{Computer Science Department - Rutgers, the State
University of New Jersey, NJ, USA\\
E-mail: kostas.bekris@cs.rutgers.edu}}

\maketitle
\IEEEpeerreviewmaketitle

\input{notation}

\begin{abstract}
\input{00_abstract}
\end{abstract}

\section{Introduction}
\label{sec:intro}
\input{01_introduction}

\section{Related Work and Contribution}
\label{sec:background}
\input{02_background}

\section{Problem Setup and Notation}
\label{sec:problem}
\input{03_problem}

\section{Rearrangement Planning}
\label{sec:method}
\input{04_00_method}

\section{Efficient Computations for Manipulation Paths}
\label{sec:implementation}
\input{05_00_implementation}

\section{Evaluation}
\label{sec:evaluation}
\input{06_evaluation}

\section{Discussion}
\label{sec:discussion}
\input{07_discussion}

%

\end{document}

%% file: notation.tex
\newcommand{\reals}{\mathbb{R}}
\newcommand{\integers}{\mathbb{Z}}

\newcommand{\Wspace}{\mathcal{W}}
\newcommand{\Objects}{\mathcal{O}}
\newcommand{\Manip}{\mathcal{M}}
\newcommand{\nobj}{k}

\newcommand{\Pspace}{\mathcal{P}}
\newcommand{\Pstable}{\mathcal{P}^s}
\newcommand{\pose}{p}
\newcommand{\GeomObj}{\mathcal{WO}}
\newcommand{\Arrange}{\mathcal{A}}
\newcommand{\Pumped}{\mathcal{A^P}}
\newcommand{\pumpedarr}{\alpha^{\mathcal{P}}}

\newcommand{\Qspace}{\mathcal{Q}}
\newcommand{\GeomManip}{\mathcal{WM}}

\newcommand{\Tspace}{\mathbb{T}} 
\newcommand{\Xspace}{\mathbb{X}}
\newcommand{\paths}{\Pi}

\newcommand{\roadmap}{\mathcal{R}}
\newcommand{\graph}{\mathcal{G}}
\newcommand{\nodes}{\mathcal{V}}
\newcommand{\node}{\mathcal{v}}
\newcommand{\edges}{\mathcal{E}}
\newcommand{\edge}{\mathcal{e}}
\newcommand{\prmstar}{{\tt PRM$^*$}}

\newcommand{\rpg}{${\tt RPG}$}

\newcommand{\local}{\mathcal{L}}

\newcommand{\prm}{{\tt PRM}}
\newcommand{\kprmstar}{{\tt k-PRM$^*$}}
\newcommand{\rrt}{{\tt RRT}}
\newcommand{\rrtdrain}{{\tt RRT-Drain}}
\newcommand{\rrg}{{\tt RRG}}
\newcommand{\est}{{\tt EST}}
\newcommand{\rrtstar}{{\tt RRT$^*$}}
\newcommand{\srrt}{{\tt RDG}}
\newcommand{\bvp}{{\tt BVP}}
\newcommand{\rdg}{{\tt RDG}}
\newcommand{\lrg}{{\tt LRG}}
\newcommand{\alg}{{\tt ALG}}
\newcommand{\upump}{{\tt UPUMP}}
\newcommand{\prxpump}{{\tt RPG}}
\newcommand{\fixed}{{\tt Fixed}-$\alpha$-\rdg}
\newcommand{\nrob}{k}
\newcommand{\cons}{K}

\newcommand{\frnodes}{V_f}
\newcommand{\frnode}{v_f}
\newcommand{\grnodes}{V_g}
\newcommand{\grnode}{v_g}
\newcommand{\fredges}{E_f}
\newcommand{\fredge}{e_f}
\newcommand{\gredges}{E_g}
\newcommand{\gredge}{e_g}
\newcommand{\kedges}{E_{\cons}}
\newcommand{\kedge}{e_{\cons}}
\newcommand{\safe}{q_s^{\mathcal{M}}}
\newcommand{\hedges}{E_H}
\newcommand{\hedge}{e_H}
\newcommand{\hnodes}{V_H}
\newcommand{\hnode}{v_H}
\newcommand{\hgraph}{H}
\newcommand{\nblank}{b}
\newcommand{\config}{C}
\newcommand{\cquery}{\mathbb{C}}
\newcommand{\pumped}{P}
\newcommand{\pumpedgraph}{\mathcal{G}_P}
\newcommand{\pnodes}{V_P}
\newcommand{\pnode}{v_P}
\newcommand{\pedges}{E_P}
\newcommand{\pedge}{e_P}
\newcommand{\signs}{\Sigma}
\newcommand{\sign}{\sigma}
\newcommand{\gsign}{\sigma_{\pumpedgraph}}
\newcommand{\cedges}{E_c}
\newcommand{\constraints}{\tt c}

\newenvironment{myitem}{\begin{list}{$\bullet$}
{\setlength{\itemsep}{-0pt}
\setlength{\topsep}{0pt}
\setlength{\labelwidth}{0pt}
\setlength{\leftmargin}{10pt}
\setlength{\parsep}{-0pt}
\setlength{\itemsep}{0pt}
\setlength{\partopsep}{0pt}}}%
{\end{list}}

%% file: 00_abstract.tex
This work proposes a method for effectively computing manipulation
paths to rearrange similar objects in a cluttered space.  The solution
can be used to place similar products in a factory floor in a
desirable arrangement or for retrieving a particular object from a
shelf blocked by similarly sized objects. These are challenging
problems as they involve combinatorially large, continuous
configuration spaces due to the presence of multiple moving bodies and
kinematically complex manipulators. This work leverages ideas from
algorithmic theory, multi-robot motion planning and manipulation
planning to propose appropriate graphical representations for this
challenge. These representations allow to quickly reason whether
manipulation paths allow the transition between entire sets of objects
arrangements without having to explicitly enumerate the path for each
pair of arrangements. The proposed method also allows to take
advantage of precomputation given a manipulation roadmap for
transferring a single object in the same cluttered space. The resulting
approach is evaluated in simulation for a realistic model of a ${\tt
Baxter}$ robot and executed in open-loop on the real system, showing
that the approach solves complex instances and is promising in terms
of scalability and success ratio.

%% file: 01_introduction.tex
Robot manipulators can benefit from the ability to rearrange objects
in constrained, cluttered human environments.  Such a skill can be
useful, for instance, in manufacturing, where multiple products need
to be arranged in an orderly manner or in service robotics where a
robotic assistant, in order to retrieve a refreshment from a
refrigerator, must first rearrange other items. This paper proposes a
methodology for solving such tasks in geometrically complex and
constrained scenes using a robotic arm. The focus is on the case that
the target objects are geometrically similar and interchangeable.

A key challenge in developing practical algorithms for rearrangement
challenges is the size of the search space. A complete method must
operate in the Cartesian product of the configuration spaces of all
the objects and the robot. Problems also become hard when the objects
are placed in tight spaces, coupled with limited manipulator
maneuverability. This paper deals primarily with these combinatorial
and geometric aspects and proposes motion planning methods that return
collision-free paths for manipulating rigid bodies. Several issues
also arise in real-world applications, which are studied here, such as
accurate estimation of object locations from sensing data.
   
The approach reduces the continuous space, high-dimensional
rearrangement problem into several, discrete challenges on
``rearrangement pebble graphs'' (\rpg s).  The inspiration comes from
work in algorithmic theory on ``pebble graphs'' \cite{Auletta:1999hc},
recent contributions in multi-robot motion
planning \cite{Solovey2012k-Color-Multi-R}, as well as work in
manipulation planning \cite{Simeon:2004tg}.


\begin{figure}
\centering
\includegraphics[width=0.45\textwidth]{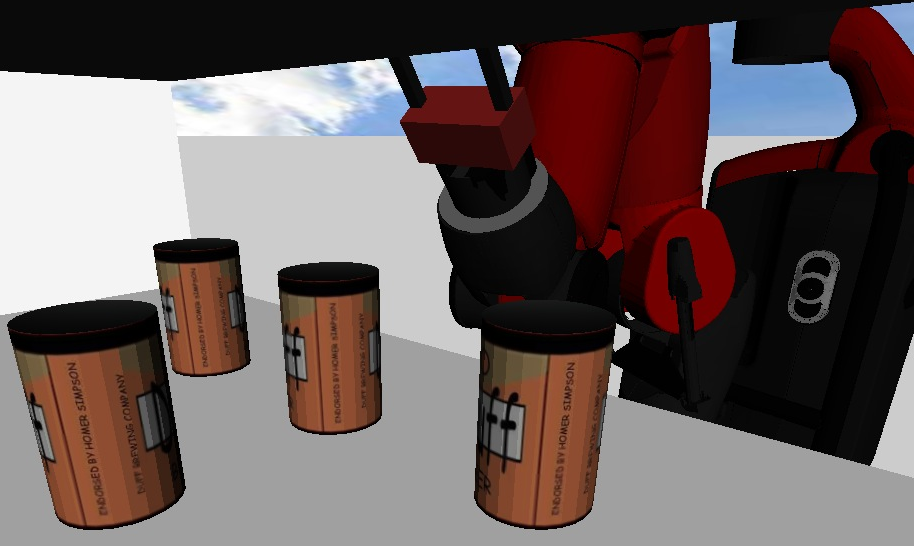}

\caption{In order for {\tt Baxter} to grasp
  the can at the back of the shelf, the other cans need to be
  rearranged.}
\label{fig:intro}
\end{figure}

The \rpg s contain stable poses for the objects as nodes.  Edges
indicate that there is a path for which the manipulator can transfer
an object between the corresponding poses, even if all other poses are
occupied.  For most interesting queries, it is difficult to find a
single pebble graph that contains both the start and the goal object
arrangements and solves the problem. Instead, it is necessary to
generate multiple such graphs and then identify transitions between
them to find a solution. \rpg s can implicitly represent an entire set
of object arrangements over their nodes. Given that the objects are
similar and unlabeled, objects can be safely moved along an edge of
an \rpg. Only once a sequence of pebble graph transitions that solves
the problem has been found does the algorithm need to extract an
explicit sequence of object placements and the corresponding
manipulation paths.  This encapsulation of multiple object
arrangements within an individual graph helps with the combinatorial
aspects of rearrangement planning. Furthermore, the method can
effectively utilize precomputation. For a known workspace and geometry
for the objects, it is possible to precompute a manipulation graph and
perform many expensive collision checking operations offline.

Relative to the state-of-the-art, the approach does not drastically
limit the type of rearrangement challenges that can be addressed to
achieve efficiency. Nevertheless, it does make certain concessions.
For instance, it must be possible to retract the arm to a safe
configuration from every stable grasped pose in a solution sequence.
Given this requirement for solutions, probabilistic completeness can
be argued withiin this set while also achieving computational
efficiency, the same way that it was achieved for multi-robot path
planning \cite{Solovey2012k-Color-Multi-R}. Furthermore, the relation to this
previous work implies a way to extend the proposed framework to
dissimilar objects. Nevertheless, the transfer of the idea of pebble
graphs to the problem of rearrangement was not trivial.  The presence
of a manipulator in rearrangement planning induces additional
constraints relative to the previous work on multi-robot motion
planning \cite{Solovey2012k-Color-Multi-R}, which required the development of
different solutions for the connection of \rpg s.

%% file: 02_background.tex
Rearrangement planning \cite{Ben-Shahar:1998pi, Ota:2004mi} has been
approached from many different perspectives in the robotics
literature.

{\bf Planning amove Movable Obstacles:} A related challenge for mobile
systems is the problem of navigation among movable obstacles
(NAMO). Early on it was shown that NAMO is
NP-hard \cite{Wilfong:1988ys} and later on it was confirmed that this
is the case even for simpler instances that involve only unit square
obstacles \cite{Demaine:2000kl}. Due to the problem's complexity, most
efforts have focused on efficiency \cite{Chen:1991fk,
Nieuwenhuisen:2006nx} and provide completeness only in certain
cases \cite{Stilman:2004qa, Stilman:2006oq}. A probabilistically
complete solution for NAMO was eventually provided \cite{Berg:2008vn},
but it can only be applied to lower dimensional robots (2-3 DOFS) and
corresponds to uninformed brute force search. More recently, a
decision-theoretic framework for NAMO has also been presented, which
deals with the inherence uncertainty in both perception and control of
real robots \cite{Levinh:2012kx}.

An interesting related challenge is the minimum constaint removal
problem, where the goal is to minimize the amount by which constraints
must be displaced to get a feasible path \cite{Hauser:2012ly}. For
this problem any asymptocally optimal solution has recently been
achieved \cite{Hauser2013Minimum-Constraint}but the problem does not
consider negative interactions between obstacles as in the current
setup.

{\bf Manipulation Planning and Grasping:} The focus in this work is on
building solutions for high-DOF robotic arms and solve manipulation
challenges. Such problems can be approached with a multi-modal
``manipulation graph'' abstraction that contains ``transit'' and
``transfer'' paths \cite{Alami:1989uq, Alami:1997fk, Simeon:2004tg}
and which can be constructed through a sampling-based
process \cite{Kavraki1996Probabilistic-R, LaValle2001}. This paper
follows a similar formalization and applies it appropriately to the
case of multiple similar movable objects towards achieving an
efficient solution. The current work is also utilizing asymptotically
optimal sampling-based planners for the computation of the
manipulation graph \cite{Karaman2011Sampling-based-}. Tree-based
sampling-based planners have also been used successfully in the
context of manipulation planning \cite{Berenson:2009hc,
Berenson:2012ij}.

There is also a variety of different approaches for manipulation
planning, which employ heuristic search \cite{Cohen:2010kx} or
optimization methodologies, such as CHOMP \cite{Zucker:2013vn,
King:2013bs}. The focus of the current paper is more on the
combinatorial challenges that arise from reasoning about multiple
objects and not the actual manipulation method. In this process, this
work is making significant use of heuristic search primitives over
sampling-based roadmaps. The output of the current algorithm could
also potentially be integrated with methods like CHOMP to improve the
quality of the computed solution. There is also significant effort
that focuses on identifying appropriate grasps in complex
scenes \cite{Berenson:2007ly, Ciocarlie:2010fu} but the current paper
does not focus on this aspect of the challenge.

The above manipulation efforts deal with moving and grasping
individual objects. Manipulation planning among multiple movable
obstacles has been considered before for ``monotone'' problems where
each obstacle can be moved at most once \cite{Stilman:2007kl,
Mike:2013fv}.  The solution in the current paper can reason about more
complex challenges. Assembly planning is also solving similar
multi-body problems but the focus there is on separating a collection
of parts and typically the robot path is ignored \cite{Wilson:1994fk,
Halperin:2000uq, Sundaram2001Disassembly-Seq}. Another paradigm for
dealing with cluttered scenes involves non-prehensile manipulation,
such as pushing objects \cite{Cosgun:2011cr, Dogar:2011ve}. The
current solution could potentially be extended to include such actions
but the focus in this paper is on grasping primitives.

{\bf Task and Motion Planning:} Rearrangement planning is an instance
of integrated task and motion planning, which can be seen as an
important step towards solving complex challenges in robotics and
especially in manipulation \cite{Cambon2009A-Hybrid-Approach-,
Hauser2011Randomized-Multi-Modal-,
Kaelbling2012Integrated-Robot}. Such integrated problems have been
approached through multi-modal roadmap structures before to deal with
the presence of both discrete and continuous
parameters \cite{Bretl:2004ys,Hauser:2010zr,
Hauser2011Randomized-Multi-Modal-}, which is also useful in the
current challenge.

{\bf Multi-robot Motion Planning:} A major motivation for the current
work is to utilize recent progress in multi-robot motion planning
solvers \cite{Berg:2009ve, Luna:2011tg, Wagner:2012cr, Yu:2012dq} in
the context of manipulation planning.  Multi-robot motion planning is
itself a hard problem due to the high-dimensionality introduced from
the presence of multiple moving bodies. Thus, coupled, complete
approaches typically do not scale well with an increased number of
robots, although there are efforts in decreasing the number of
effective degrees of freedom \cite{Aronov:1999nx}. On the other hand,
decoupled methods, such as priority-based schemes \cite{Berg:2005bh}
or velocity tuning \cite{Leroy:1999qf}, trade completeness for
efficiency.

An interesting line of work in algorithmic theory focuses on ``pebble
motion on graphs'', where disticnt pebbles need to move from an
initial to a goal vertex assignment on a
graph \cite{Kornhauser:1984oq}. The unlabeled version of the problem
has also been studied in the literature \cite{Calinescu:2008kl}. It
turns out that testing the feasibility of a ``pebble motion'' problem
can be answered in linear time \cite{Auletta:1999hc, Goraly:2010ij}.
These results from algorithmic theory, integrated with sampling-based
algorithms for motion planning, inspired a recent method for in
continuous representations \cite{Solovey2012k-Color-Multi-R}. This
approach reduces the multi-robot motion planning problem into a
sequence of discrete pebble problems in a manner that is possible for
the algorithm to efficiently transform movements of pebbles into valid
motions of the robots.

The current paper is motivated by the progress achieved by this recent
multi-robot motion planning work and the results in algorithmic
theory. It aims to take advantge of similar primitives from the
``pebble motion'' problem to solve in rearrangement planning
challenges where objects motions have to also respect manipulation
constraints.

%% file: 03_problem.tex
\begin{figure}
\begin{centering}
\includegraphics[width=0.24\textwidth]{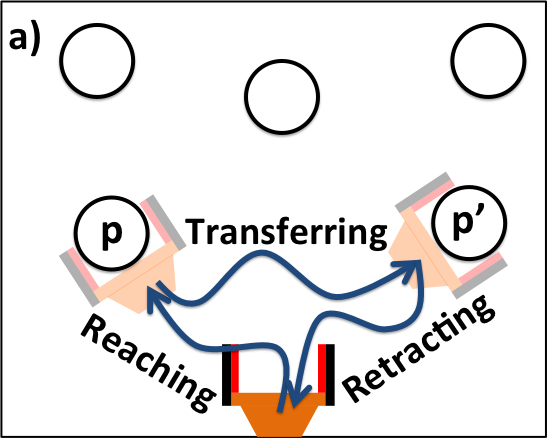}
\includegraphics[width=0.24\textwidth]{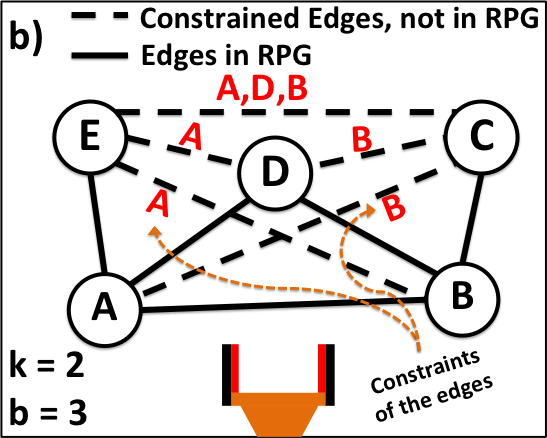}
\caption{Each edge on an \rpg\ is the combination of a reaching, transferring and retracting path. }
\label{fig:rpg}
\end{centering}
\end{figure} 

Consider a {\bf workspace} $\Wspace \subset \reals^3$ with static
obstacles, a set of $\nobj$ movable unlabeled rigid body objects with
the same geometry, and a manipulator $\Manip$. The subset of $\Wspace$
not occupied by static obstacles is defined as $\Wspace^{f}$. Each
{\bf object} $\Objects_i$ can acquire a {\bf pose} $p_i
\in \Pspace$ and for that pose its geometry occupies a subset of the
workspace. The poses are general 3D transformations, i.e.,
$\Pspace \subseteq SE(3)$. For {\it collision-free} poses
$p_i^f \in \Pspace^{f}$, the interior of the objects are placed in the
collision-free workspace. This definition allows the objects to touch
obstacles.

An {\bf unlabeled arrangement} $\alpha \in \Arrange$ specifies a
$k$-combination of poses $\{ p_1, \ldots, p_{\nobj} \}$, where
$p_i \in \Pspace$ and results in a placement of the objects in
$\Wspace$.  Permutations of $\alpha$ are indistinguishable, since the
objects are unlabeled. A {\it collision-free arrangement} $\alpha^f
\in \Arrange^f$ does not cause collisions among objects, as well as
between obstacles and objects. Similarly define {\bf configurations}
$q^{\Manip} \in \Qspace^{\Manip}$ for the {\bf manipulator} $\Manip$.
{\it Collision-free configurations} $q^f$ place the manipulator in the
collision-free workspace $\Wspace^{f}$.


The {\bf configuration space} $\Qspace
 = \Qspace^{\Manip} \times \Arrange$ corresponds to the Cartesian
 product of the manipulator's c-space and the space of unlabeled
 object arrangements. The {\it collision-free configuration space}
 $\Qspace^f$ does not allow collisions between the manipulator or the
 objects with obstacles, as well as between objects, or between the
 manipulator and objects. While penetrations are not allowed between
 objects and the robot, the arm can grasp the objects.  

The manipulation problem has then two types of valid configurations:
(a) {\bf Stable configurations} $q^{s} \in \Qspace^{s}$:
Collision-free configurations where the objects are in stable poses
where they rest without any forces from the arm; (b) {\bf Grasping
configurations} $q^{g} \in \Qspace^{g}$: These correspond to
configurations where the manipulator is grasping an object.  Let
$\Qspace^{v} = \Qspace^{s} \cup \Qspace^{g}$ denote the {\it valid
configuration space} for the rearrangement problem. The simulations
for this paper use as the stable set of poses
$\Pstable \subset \Pspace$ the placement of the objects on horizontal
surfaces, i.e., $\Pstable \equiv SE(2)$.

%


Furthermore, define a {\bf type} $t \in \Tspace$ for such manipulation
problems (also called a ``mode''), which has two values.  In the {\it
transit} type, the manipulator is not carrying an object and
$q \in \Qspace^{s}$. In the {\it transfer} type, the manipulator
carries an object and $q \in \Qspace^{g}$. The only way that the type
can change is if $q \in \Qspace^{t}
= \Qspace^{s} \cap \Qspace^{g}$. Such a configuration $q$ is called a
{\it transition} configuration. The problem type can change only if
the configuration of the problem is a transition one.

The problem's {\bf state space} $\Xspace: \Qspace \times \Tspace$
arises from the combination of the configuration space $\Qspace$ and
the type space $\Tspace$.  It is then easy to define the sets
$\Xspace^{f}$ (free), $\Xspace^{s}$ (stable), $\Xspace^{g}$ (grasping)
and $\Xspace^{t}$ (transition), e.g., ($\Xspace^{t}
= \Qspace^{t} \times \Tspace$).

{\bf The Unlabeled Rearrangement Problem:} Given an initial state $x_0
= ( (q_0, \alpha_0), t_0) \in \Xspace^{v}$ and a final state $x_1 = (
(q_1, \alpha_1), t_1) \in \Xspace^{v}$, compute a continuous {\it
path} $\pi \in \paths: [0,1] \rightarrow \Xspace^{v}$ such that $\pi(0) =
x_0$, $\pi(1) = x_1$ and the types of states along path $\pi$ change
only when $\pi(s) \in \Xspace^{t}$. 

Thus, a path is an alternating sequence of transit and transfer
states, which change at transition configurations. Additional
definitions and notation are introduced as needed during the
description of the proposed solution.

%% file: 04_00_method.tex
One way of solving the unlabeled rearrangement problem would be to
build a manipulation graph \cite{Alami:1997fk} in the entire
$\Xspace$. This is, however, a high-dimensional space and given that
motion planning is hard, efficient solutions cannot be achieved easily
as the number of objects increases. The idea here is to abstract out
the motion of the manipulator and then reason directly about the
movement of objects between different stable poses in
$\Pstable$. Reasoning about the movement of multiple objects can take
place over discrete graphical representations so as to take advantage
of linear-time path planning tools for rearranging unlabeled
``pebbles'' on a graph from an initial to a target
arrangement \cite{Auletta:1999hc}.

A sampling approach can be used to define graphs where nodes
correspond to stable poses in $\Pstable$ and edges correspond to
collision free motions of the arm that transfer an object between
stable poses. If such a graph is connected and contains all the poses
from the initial and target arrangements, then a discrete solver can
be used to define a solution in the continuous space as long as
placing objects in different poses does not cause
collisions \cite{Auletta:1999hc}.

It may be difficult or impossible, however, to construct a single such
graph that directly solves the problem. For example, the poses in the
initial and target arrangements could be already overlapping, or it
may not be possible to ensure connectivity with collision-free motions
of the arm. Motivated by work in the multi-robot motion planning
literature \cite{Solovey2012k-Color-Multi-R}, the current paper
considers multiple such graphs, referred to ``rearrangement pebble
graphs'' (\rpg s).

Within each \rpg\ the discrete solver can be used to achieve all
feasible arrangements, given its connectivity. If the \rpg\ has one
connected component, then all possible arrangements over the graph can
be attained for unlabeled objects and they do not need to be
explicitly stored. If the \rpg\ has multiple connected components, a
signature, which specifies how many objects exist in each connected
component, is sufficient to describe the feasible arrangements. It
should also be possible to switch between different \rpg s, if they
share at least $\nobj$ poses that can be occupied by objects, given
the corresponding signatures. 

This gives rise to a hyper-graph structure, where each node
corresponds to an \rpg\ and a signature. Edges correspond to
transitions between such hyper-nodes. The initial and target
arrangements define two such hyper-nodes. Then the approach generates
and connects hyper-nodes until the initial and target arrangement are
connected on the hyper-graph. At that point, the rearrangement problem
is solved and the necessary motions of the manipulator can be
extracted along the path connecting the initial and target nodes on
the hyper-graph.

\subsection{Rearrangement Pebble Graphs}
\input{04_01_create_pg}

\subsection{Constructing Hypernodes}
\input{04_02_create_hypernode}

\subsection{Connecting Hypernodes}
\input{04_03_connection}

\subsection{Creating The Hypergraph}
\input{04_04_create_hypergraph}

\subsection{Answering Queries}

\input{04_05_query}

\subsection{Smoothing Solutions}
\input{04_06_smoothing}

\section{Properties}
\input{05_properties}

%% file: 04_01_create_pg.tex
\rpg s are constructed so that the objects are placed in collision-free,
stable poses. They contain at least $\nobj$ poses as nodes and $b$
additional nodes. The extra nodes allow the rearrangement of $\nobj$
objects on the graphs. The set of poses used in an \rpg\ is defined as
a pumped arrangement:

\noindent \textit{Definition:} A pumped arrangement $\pumpedarr$ is a
set of $n=\nobj + \nblank$ poses where: (a) These poses are a subset
of collision-free, stable poses $\Pstable$. (b) No two objects would
collide if they were placed on any two poses of the pumped
arrangement.

\noindent \textit{Definition:} An \rpg\ is a graph
$\pumpedgraph(\pumpedarr, \edges^\pumped)$ where the set of nodes
corresponds to a pumped arrangement $\pumpedarr$. The set of edges
$\edges^\pumped$ corresponds to pairs $p, p' \in \pumpedarr$, for
which the manipulator can transfer objects between poses $p, p'$
without collisions given that potentially every other pose of
$\pumpedarr$ is occupied.

\begin{algorithm}[ht]
\caption{${\tt CREATE\_RPG} (\safe, \nrob, \nblank)$}
\label{algo:create_rpg}
$\pnodes \gets {\tt Sample\_Valid\_Pumped\_Arrangement}(\nrob
+ \nblank )$ \tcp*{sample $\nrob+\nblank$ poses}
$\pedges \gets \emptyset$, $\cedges \gets \emptyset$ \tcp*{initialize
data structure}
\For{$\pose, \pose' \in \pnodes$ and $\pose \neq \pose'$}
{
    $\pi \gets {\tt Compute\_Minimum\_Conflict\_Path}(\safe, \pose, \pose' )$\;
    \If{${\tt is\_valid}(\pi)$}
    {
        $\pedges \gets \pedges \cup ((\pose,\pose'),\pi)$ \tcp*{edge is added to the \rpg}
    }
    \Else
    {
    	$\constraints \gets Compute\_Constraints(\pi)$ \tcp*{find the constraining poses}
    	$\cedges \gets \cedges \cup ((\pose,\pose'),\pi, \constraints)$ \tcp*{store constrained edge in $\cedges$}
    }
}
\Return \{$\pumpedgraph(\pnodes, \pedges), \cedges$\}
\end{algorithm}

Algorithm \ref{algo:create_rpg} describes the construction of an
an \rpg. The algorithm receives as input a ``safe'' configuration of
the manipulator $\safe$, which is a configuration that does not
interfere with the objects placed on any stable poses and typically is
a retracted arm configuration. The algorithm needs also the size of
the \rpg, i.e., the parameters $\nobj$ and $\nblank$. The algorithm
will return an \rpg\ and a set of edges $\cedges$ that were not added
to the \rpg\ together with the poses that blocked them.

The algorithm starts by selecting a non-intersecting set of $\nrob
+ \nblank$ poses that create a pumped arrangement (Line 1). For each
pair of poses $\pose$ and $\pose'$ a path has to be computed that
allows the manipulator to move an object from $\pose$ to
$\pose'$. This path consists of three segments as shown in
Fig. \ref{fig:rpg}:

\begin{myitem}
\item[i)] A path for the arm from
$\safe$ to a transition state, $x^{t}_{\pose} \in \Xspace^{t}$, for
pose $\pose$, 
\item[ii)] A path that transfers the object from
$x^{t}_{\pose} \in \Xspace^{t}$ to a state
$x^{t}_{\pose'} \in \Xspace^{t}$ for $\pose'$,
\item[iii)]  A retraction path from $x^{t}_{\pose'}$ back to $\safe$.
\end{myitem}

\noindent For each segment the algorithm aims to compute
a ``minimum conflict path'' given that objects may be placed in all
poses but $\pose$ and $\pose'$. If there is a collision-free path,
then an edge ($\pose$,$\pose'$) is added to the \rpg\ (Line 6). The
path used to validate the transition is also stored on the edge. In
case the ``minimum conflict path'' $\pi$ collides with objects on
other poses of the \rpg, the edge ($\pose$,$\pose'$) is not added. The
algorithm identifies these potential collisions and stores them in a
set of constraints $\constraints$ (Line 8). Edges with constraints are
stored in the data structure $\cedges$ (Line 9). The algorithm returns
the \rpg\ and the set $\cedges$ of constrained edges. The use of
$\cedges$ is described later on.

%% file: 04_02_create_hypernode.tex
Objects within each connected component can be rearranged, since the
edges within an \rpg\ correspond to valid motions, regardless of
object placement. The number of objects actually in each connected
component is tracked by a signature.

\noindent \textit{Definition:} A signature, $\gsign(\alpha)$ is the
number of objects contained in each connected component of \rpg\
$\pumpedgraph$ according to an arrangement $\alpha$.

The important observation is that all arrangements in an \rpg\ which have the same signature are reachable one from another, for unlabeled objects. That is, if $\gsign(\alpha) = \gsign(\alpha')$, then there is some sequence of transitions and a corresponding path for the manipulator, $\pi$, which brings $\alpha$ to $\alpha'$, and vice versa. 

\noindent \textit{Definition:} A hyper-node is an \rpg\ and a signature $\sign$\ shared by a set of reachable arrangements of objects on the \rpg.

\begin{algorithm}[ht]
\caption{${\tt CREATE\_HYPERNODES} (\hgraph(\hnodes,\hedges), \safe, \nrob, \nblank)$}
\label{algo:create_hypernode}
$\{\pumpedgraph, \cedges \} \gets {\tt
CREATE\_RPG}(\safe, \nrob, \nblank)$ \tcp*{create an \rpg}
$\signs \gets {\tt Generate\_Signatures}(\pumpedgraph)$ \tcp*{compute
all signatures}
$V_n \gets \emptyset$ \tcp*{initialize \textquotedblright
set of sibling \rpg s}
\For{$\sign \in \signs$}
{
	$\hnode \gets (\pumpedgraph, \sign)$ \tcp*{construct a new hyper-node}
    $V_n \gets V_n \cup \hnode$ \tcp*{keep the new node as a \textquotedblright sibling\textquotedblright\ node}
    $\hnodes \gets \hnodes \cup \hnode$ \tcp*{add the new node to the hyper-graph}
    ${\tt  CONNECT\_NODE}(\hgraph, \hnode )$ \tcp*{connect new node to non-siblings}
}
${\tt CONNECT\_SIBLINGS}(\hgraph, V_n, \cedges )$ \tcp*{connect \textquotedblright siblings\textquotedblright}
\end{algorithm}


Algorithm \ref{algo:create_hypernode} takes as an argument the
existing hyper-graph and adds new hyper-nodes and edges. It needs to
be aware of the safe configuration $\safe$, and the size of the \rpg s
it will construct: $n = \nobj + \nblank$. When this function
completes, the new nodes will be in the hyper-graph and appropriately
connected with edges.

The algorithm begins by constructing a random, valid \rpg\ (Line 1).  Then, it computes all possible signatures that the generated \rpg\ can attain (Line 2). A new hyper-node $\hnode$ is created for each signature (Line 5). When $\hnode$ is added in the hyper-graph $\hgraph$, function ${\tt CONNECT\_NODE}$ tries to connect $\hnode$ with the already existing nodes in $\hgraph$ corresponding to different \rpg s (Line 8). Finally, when all the ``siblings'', nodes that correspond to  the same \rpg\ have been created, function ${\tt CONNECT\_SIBLINGS}$ will try to add edges between them (Line 9).

%% file: 04_03_connection.tex
 
\begin{figure}
\centering
\includegraphics[width=1.72in]{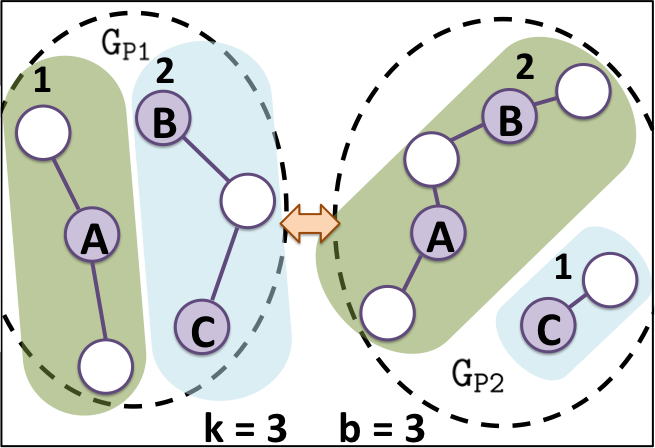}
\includegraphics[width=1.72in]{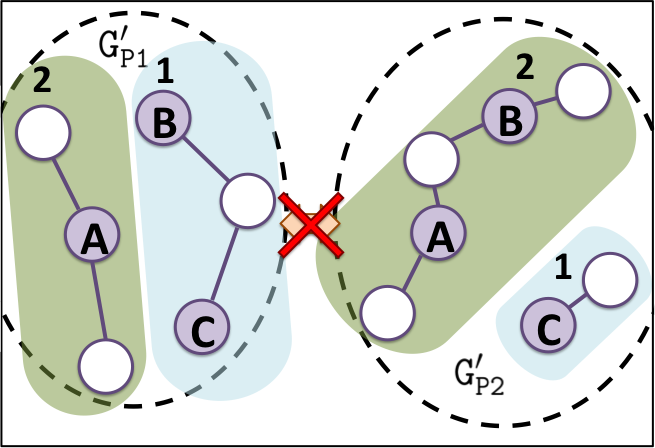}
\caption{An illustration of ${\tt CONNECT\_NODE}$ that both \rpg s
  share the shaded positions (A,B,C). (Left) A connection is made
  between the nodes, as the signature information allows all three
  common positions to contain an object in both \rpg s.  (Right) The
  two nodes cannot be connected, because on $\mathcal{G}'_{P1}$ it is
  not possible to place objects on both common poses B and
  C.}
\label{fig:hyperedges}
\end{figure}

\begin{figure}
\centering
\includegraphics[width=1.72in]{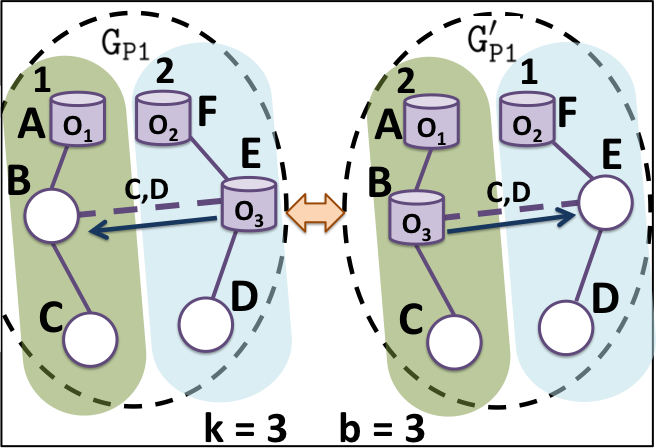}
\includegraphics[width=1.72in]{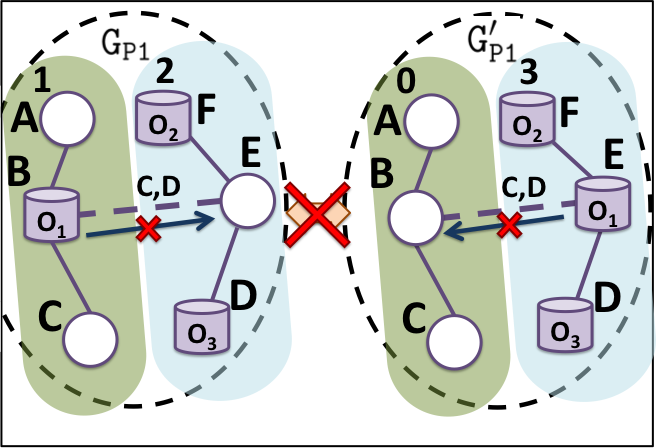}
\caption{${\tt CONNECT\_SIBLINGS}$: Nodes from the same \rpg but
  different signatures. The constrained (dashed) edge is not part of
  the \rpg\ but can be used to change signature. (Left) The nodes can
  be connected as moving an object along the edge is feasible given
  the constraints (C,D).  (Right) The two nodes cannot be connected,
  because the pose D cannot be emptied. }
\label{fig:siblingsedges}
\end{figure}

The hyper-nodes in $\hgraph$ have two different ways to connect to
each other: (A) ${\tt CONNECT\_NODES}$ connects nodes from a new
\rpg\ with nodes from previous \rpg s, based on similar poses and
signatures. The function first identifies whether node $\hnode$ shares
at least $\nobj$ common poses with an existing node. If true, the
method checks whether both nodes can achieve placing $\nobj$ objects
on the $\nobj$ common poses given the nodes' signatures
(Fig. \ref{fig:hyperedges}). If they both can, an edge is added
between the two nodes, which represents a context switch between two
arrangement sub-problems. (B) ${\tt CONNECT\_SIBLINGS}$ connects nodes
that come from the same \rpg\ by using the constraint edges $\cedges$,
to switch between different signatures on the same \rpg. The algorithm
connects nodes only if there is a motion that allows transition
between signatures of the same \rpg\ (Fig. \ref{fig:siblingsedges}).

The arm is able to follow a constrained edge in $\cedges$\ if the
constraining poses can be emptied given the hyper-node's
signature. For all the constrained edges the algorithm finds the
connecting components of the \rpg\ that this edge is connecting. If
the connecting components are different and the edge is feasible given
the signature, then this edge can be used for potential connections
between two hyper-nodes. An edge is feasible if the poses that
constrain the edge can be emptied given the signature of $\hnode$ and
the target pose $\pose'$ of the constrained edge can be
emptied. Furthermore, it must be possible to bring an object to the
source pose $\pose$ of the edge. Moving along such edges results in a
change of signature relative to that of $\hnode$ and the new signature
$\sign_n$ can be computed. Then an edge is created that connects the
two ``sibling'' nodes in the hyper-graph.

%% file: 04_04_create_hypergraph.tex
\begin{figure*}
\centering
\includegraphics[width=2.2in]{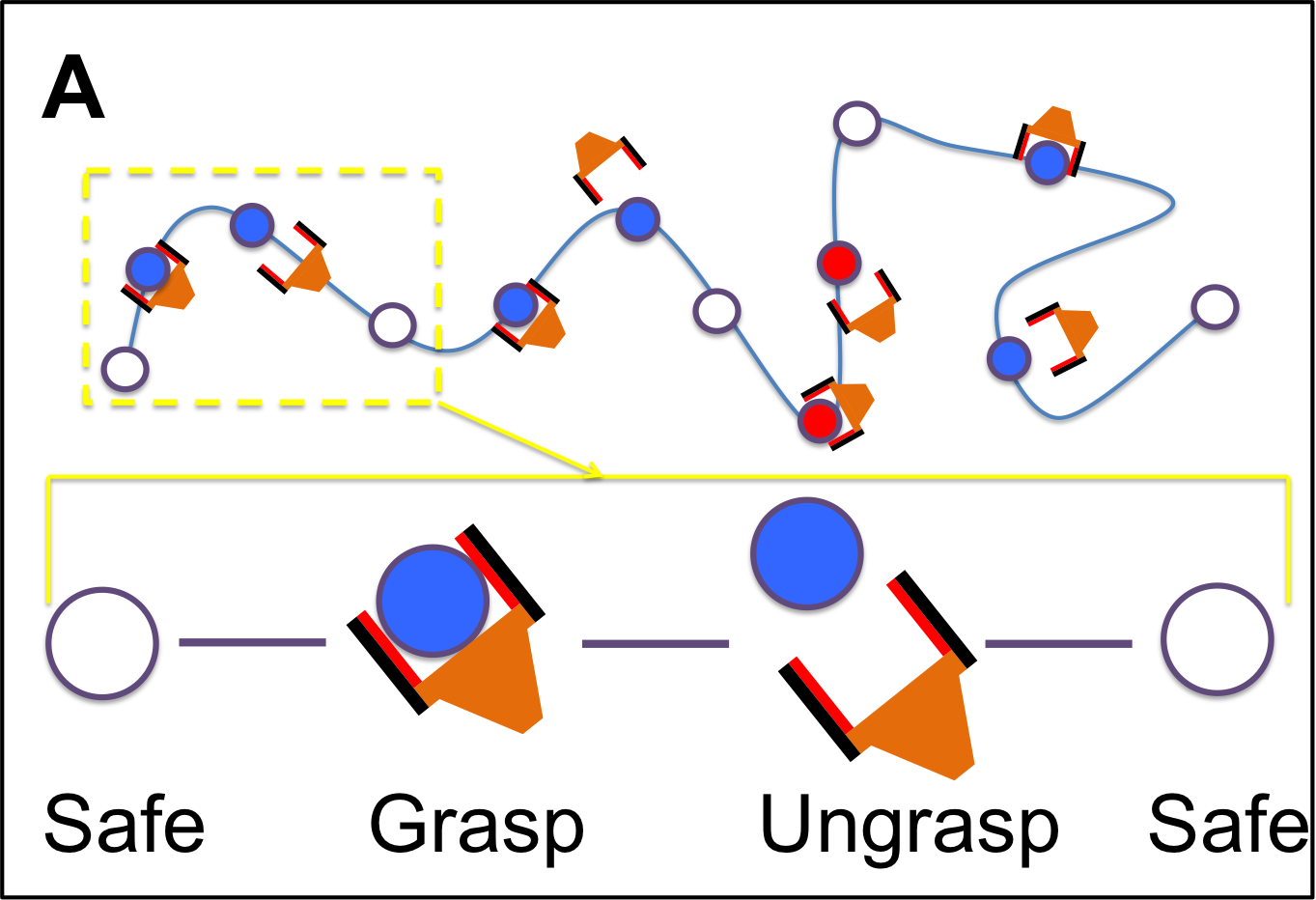}
\hspace{0.1in}
\includegraphics[width=2.2in]{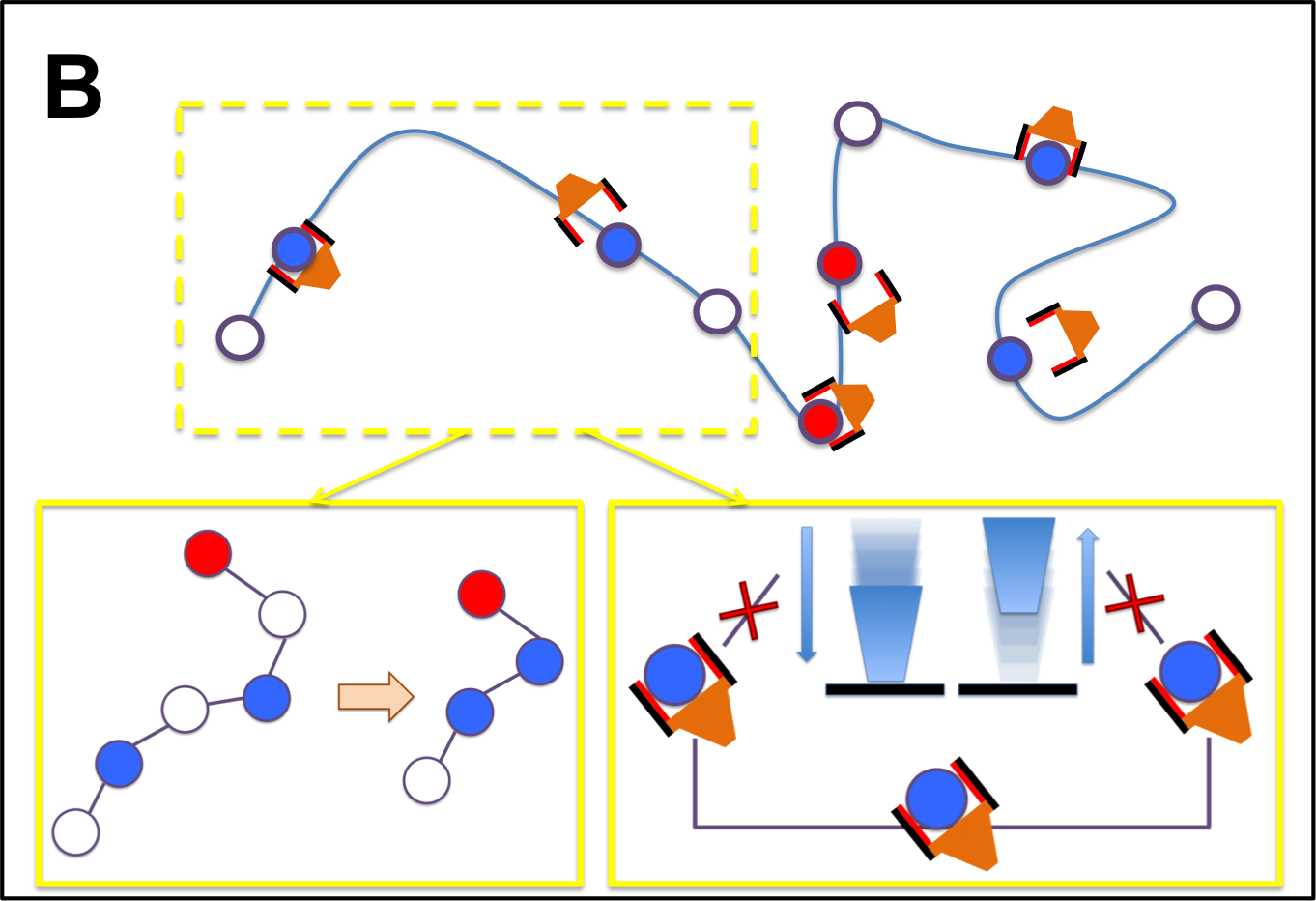}
\hspace{0.1in}
\includegraphics[width=2.2in]{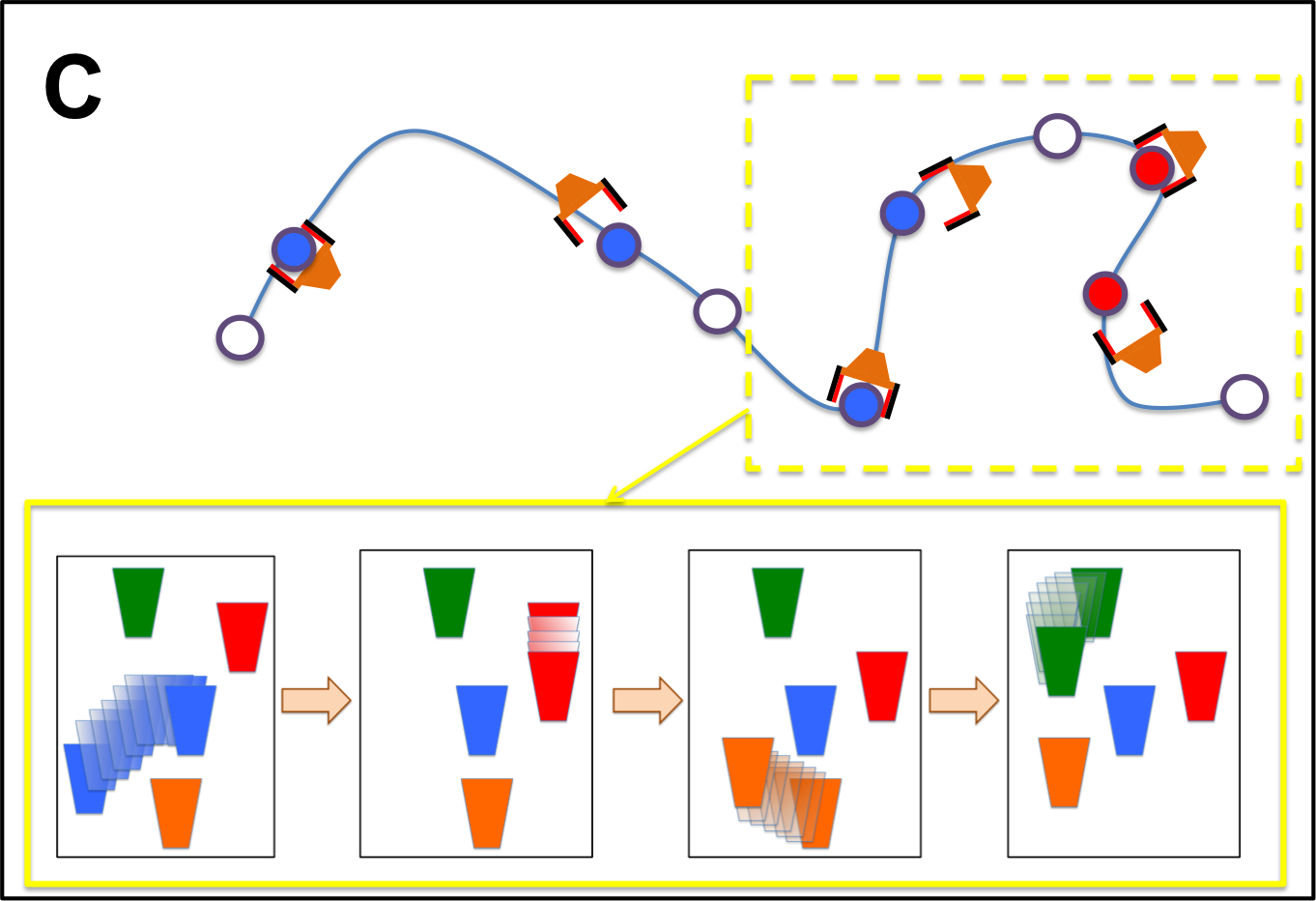}
\caption{A solution trajectory can be decomposed into intervals of
reaching, transferring and retracting as in A. If the same grasp on
the same object exists consecutively in the solution, then the
intermediate safe state is removed. In addition, any symmetrical
sequence of states, such as lowering and raising the object, can also
be removed, as in B. In order to remove redundant intermediate
placements of the same object, placement of the same object in
different parts of the trajectory are checked if they can be removed
and the object directly placed the the next position.}
\label{fig:smoothing}
\end{figure*}

The overall approach operates similarly to a bidirectional tree
sampling-based planner (e.g., $\est$) \cite{Hsu2002}.  It first
generates two hyper-nodes of size $\nobj$ for the initial and target
arrangements.  To help the generation of edges between hyper-nodes, a
seed arrangement, $\alpha_{seed}$, is used to construct the pebble
graph for new nodes in Alg. \ref{algo:create_hypernode}. It selects an
existing node on the tree randomly and a seed $\alpha_{seed}$ of
$\nobj$ poses are selected that agree with the signature of the
selected node. This seed is used inside
Alg.\ref{algo:create_hypernode} to generate the new \rpg\ and the
corresponding hyper-nodes. The method keeps doing this until the
initial and goal hyper-nodes are in the same connected component.
Upon completion, the method generates the sequence of motion plans
from the hyper-graph that satisfies the query.

%% file: 04_05_query.tex
For each hyper-node, there exists a way to move between all possible
arrangements of objects on the corresponding \rpg\ that have the same
signature.  But these actual paths that accomplish these rearrangement
are not explicitly stored. Only during the query phase, when a graph
search returns a solution sequence of hyper-nodes and edges the
algorithm computes solutions to the pebble motion problem within
each \rpg. 

The hyper-graph path is transformed into a manipulation path by
solving discrete graph problems \cite{Auletta:1999hc} on
individual \rpg s given start and final arrangements on them according
to information stored on the hyper-edges. The algorithm requires a
start and a final arrangement on the \rpg. Because the \rpg\ contains
edges which have safe motions for the manipulator, any arrangement
produced by the approach will be feasible for the robotic arm if it
follows the sequence of actions that was used to validate the
edge. The start arrangement on the \rpg\ is simply the latest
arrangement the objects have reached in the previous \rpg. The end
arrangement is stored on the edges of the hyper-graph.  For
connections between nodes generated from the same \rpg, the edges
involve a motion and have the constraints associated with this motion
encoded. The end configuration is such that those constraints are
satisfied.

%% file: 04_06_smoothing.tex
Figure \ref{fig:smoothing} shows the operation of smoothing. The colored disks represent poses occupied by an object, while the uncolored discs represent the safe configuration $\safe$. The edges connecting the nodes represent the trajectory taken by the manipulator to move the objects. A trajectory can be separated into distinct sequences of transit to grasp from $\safe$, transfer an object between poses ($\pose$,$\pose'$), and retract to $\safe$, as shown in Figure \ref{fig:smoothing}a.

\noindent\textit{Phase 1, Consecutive, Identical Grasps: } For each of these distinct sequences, the smoothing process first looks for consecutive, identical grasps on the same object. If these consecutive sequences exist, any intermediate safe state is removed, along with any redundant states, as shown in Figure \ref{fig:smoothing}b. This has the effect of removing motions such as raising and lowering the object, which is an unnecessary motion when consecutively grasping the same object.

\noindent\textit{Phase 2, Standard Trajectory Smoothing: }
Each reaching, transferring and retracting sequence is smoothed by checking for shortcuts between pairs of states. If these shortcuts exist, they replace their corresponding intermediate trajectory.

\noindent\textit{Phase 3, Maximizing Consecutive Grasps: }
The solution returned by the algorithm is conservative. This can potentially lead to redundant movements of objects. The trajectory that moves an object $o_{x}$ from position $p$ to $p'$ can be represented as $o^{p}_{x} \rightarrow o^{p'}_x$. The trajectory sequence $o^i_a \rightarrow o^{i+1}_{a} \rightarrow o^{j}_{b} \rightarrow o^{j+1}_{b} \rightarrow o^{i+1}_{a} \rightarrow o^{i+2}_{a}$ contains an intermediate placement of $o_{a}$ at pose $i+1$, before eventually moving it to $i+2$  (Figure \ref{fig:smoothing}c). If (I) $o_{a}$ can be moved from pose $i$ to pose $i+2$ with a collision free path, given that $o_b$ is at pose $j$ and (II) the trajectory of $o^j_b \rightarrow o^{j+1}_b$ is collision free with $o_a$ at pose $i+2$ then the original trajectory can be replaced with $o^{i}_{a} \rightarrow o^{i+2}_{a} \rightarrow o^{j}_{b} \rightarrow o^{j+1}_{b}$.

%

These phases can be repeated, starting from Phase 1, to continually improve the path quality. Once the smoothing phases are complete, Phase 2 can applied over the entire trajectory, resulting in the final smoothed solution. In practice, a single-pass of this process and a final application of Phase 2 are sufficient.

%% file: 05_properties.tex
Every pose along a valid edge in the \rpg \ needs to be reachable from
$\safe$.  This requirement implies that the proposed method solves
rearrangement problems that satisfy the following
property: \emph{there is a solution path for which every stable
grasped pose along this path is reachable from $\safe$}.

The problems studied here are solved in a sequential manner, given a
single arm, where the solution sequence can be decomposed into
transfer and transit paths. Each transfer segment corresponds to
$\nobj-1$ objects remaining static and one object moved from a stable
pose $\pose$ to another $\pose'$. Such a motion can be discovered by
the proposed approach if a $k+1$-pose \rpg \ is sampled, that contains
the $\nobj-1$ poses of the static objects, as well as $\pose$ and
$\pose'$. Since the solution path contains the motion
$\pose' \rightarrow \pose'$, the \rpg \ will also be able to discover
this edge, as long as poses $\pose$ and $\pose'$ are reachable from
$\safe$. So, in the worst case using \rpg s of size $\nobj+1$, this
method needs to generate all the \rpg s that correspond to each
transfer motion of a solution trajectory. The \rpg s that represent
the solution motions can be pairwise connected, since they share
$\nobj$ poses. Thereby, if all the \rpg s corresponding to transfer
paths of the solution are sampled, the edges that correspond to the
motions within the \rpg s are guaranteed to exist and the hyper-nodes
belonging to the solution will be pairwise connected. The approach
will discover such a solution, if it exists, and hence it
is \emph{probabilistically complete} within the set of problems that
satisfy the assumption about $\safe$.

In practice, higher values of ``blank'' poses $\nblank$ that 1,
provide computational benefits as a higher number of arrangements can
be represented by a single \rpg. There is a trade-off, however, since
as $\nblank$ increases, the connectivity of the \rpg \ will be
decreasing. The experimental section evaluates this trade-off.

The path planner used for computing the paths of the manipulator is
assumed to be complete in the above discussion. The following section
provides a way to compute such paths, given a pre-computed
manipulation roadmap, implying a dependence of the overall method on
the probabilistic completeness of that approach.

%% file: 05_00_implementation.tex

\begin{figure}
\begin{centering}
\includegraphics[width=0.37\textwidth]{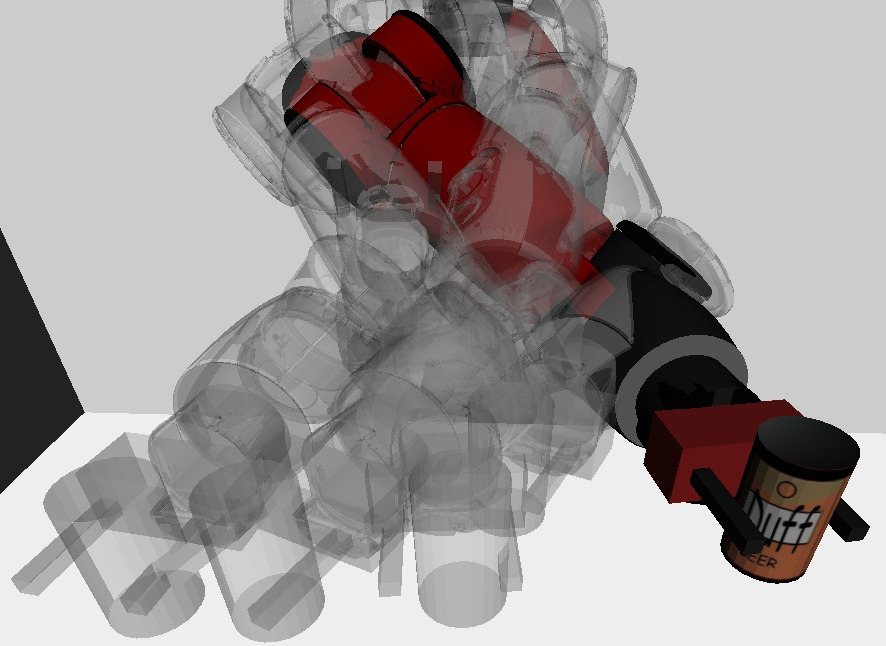}
\caption{Transition states $x^{t} \in X^{t}$, i.e., stable grasping configurations. An IK solution for the end effector is computed.}
\label{fig:transitions}
\end{centering}
\end{figure}

A manipulation planning primitive is needed during the construction of
the \rpg s to detect whether an edge on the \rpg\ can be added. For
the efficient online computation of this operation, this work
generates a manipulation roadmap $\roadmap(\nodes,\edges)$ offline,
for a single object and the static geometry. In this manner, the
roadmap operates over reduced collision-free single-object states of
the form $x = ((q^{\Manip}, p),t)$, where $q^{\Manip}
\in \Qspace^{\Manip}$ is a manipulator configuration, $p$ is the pose
of the single object and $t$ is the type, i.e., either a transfer or a
transit state. 

The first step in the construction of $\roadmap$ is to sample
transition states $x^{t} \in \Xspace^{t}$, i.e., stable grasping
configurations.  This can be achieved by first sampling a stable
collision-free object pose, $p \in \Pspace^s$, and then using inverse
kinematics to define the corresponding manipulator's grasping
configuration $q^{\Manip}$ (Fig. \ref{fig:transitions}). If the
inverse kinematics function returns a solution and the resulting
configuration $(q^{\Manip},p)$ is collision-free, then two states are
generated with the same configuration: one of the transit type and one
of the transfer type. Both of these states are inserted in $\roadmap$
and an edge is added between them.

In the experiments performed for this paper multiple attempts to
generate grasping configurations for the manipulator take place for an
individual object pose. Furthermore, two extra states are inserted in
$\roadmap$ during this process. A transit state, called a
``retraction'' state, where the arm's end-effector is slightly
retracted from the object, and a transfer state, called a ``raised''
state, where the arm's end-effector is slightly raised with the
object. These additional states help in the connectivity of both the
transit and transfer subsets of $\roadmap$.

Given the set of transition states and edges, the method proceeds to
separately explore the transit and the transfer subset of the state
space in a \prmstar\ fashion. For the transit component, additional
grasping configurations for the manipulator are sampled, where the
object is no longer required to be in a stable pose. Then neighboring
transit states according to a configuration space metric are
considered for connection by interpolating between the two
configurations. If the interpolated path is collision-free, the edge
is added to the roadmap. The same process is repeated for transfer
states. The objective is to achieve a roadmap that has the minimum
number of connected components, so that the manipulator is able to
transfer objects among the sampled transition states.

After the roadmap is constructed offline, it can also be informed
about collision between potential objects placed on stable poses and
the arm moving along edges of $\roadmap$. All the sampled poses that
have been used to build the manipulation roadmap are checked for
collisions with the edges of the roadmap. In this way, all the
necessary collision-checking can take place offline. This information
is used during the online process in order to speed up the search for
the ``minimum conflict path'' during the construction of \rpg s.

\begin{figure*}
\centering
\includegraphics[width=2.2in]{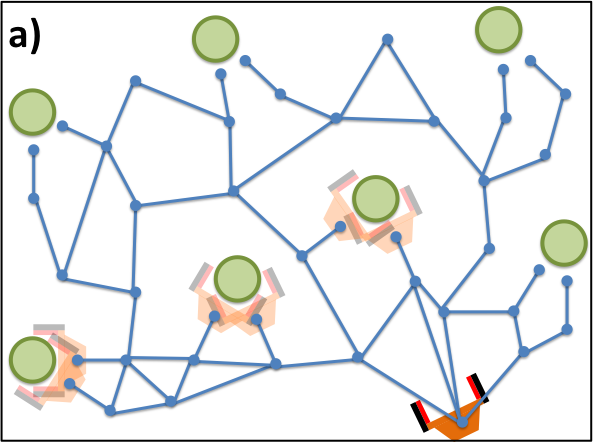}
\hspace{0.1in}
\includegraphics[width=2.2in]{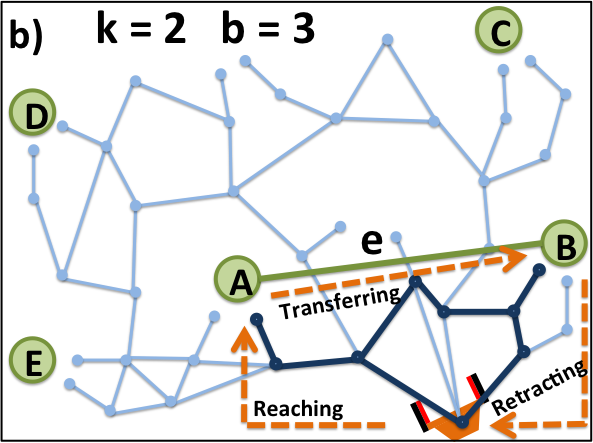}
\hspace{0.1in}
\includegraphics[width=2.2in]{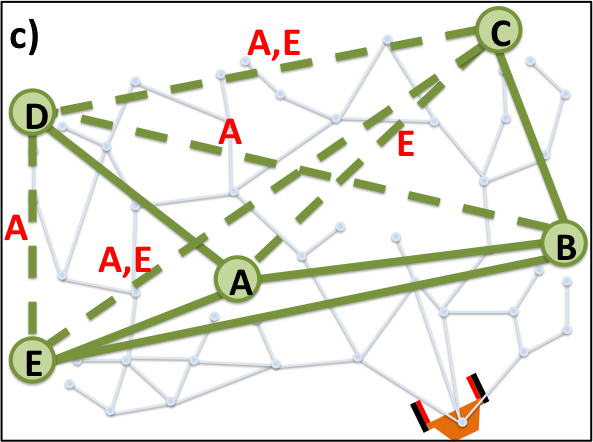}
\caption{\rpg\ creation using a manipulator roadmap $\roadmap$: a) The
  algorithm selects $\nobj + \nblank$ stable poses from $\roadmap$ to
  generates a pumped arrangement. b) An edge is added if it is
  possible given $\roadmap$ to transfer an object between these poses
  (A to B) without collisions given that all other poses are
  potentially occupied. The method identifies if the ``minimum
  conflict path'' has a constraint.  c) The new \rpg\ is returned with
  the collision free edges and the constrained edges in a separate
  structure $\cedges$. }
\label{fig:build_pumped_graph}
\end{figure*}

Fig. \ref{fig:build_pumped_graph} describes the use of the roadmap
$\roadmap$ for the online generation of \rpg s. First the poses of the
\rpg\ are selected from the poses used by $\roadmap$. The roadmap
stores multiple configurations $q^{\Manip} \in \Qspace^{\Manip}$,
which can grasp an object in a stable configuration for stable
pose. The safe configuration $\safe$ is also part of the roadmap
$\roadmap$. To add an edge to the \rpg\ between poses $\pose$ and
$\pose'$, the three segments path is evaluated on the roadmap. A
heuristic search approach for ``minimum conflict path-planning'' on
the roadmap is used for this process in order to find the least
constrained path , $\safe \rightarrow x^{t}_{\pose} \rightarrow
x^{t}_{\pose'} \rightarrow \safe$ given the other poses.

%% file: 06_evaluation.tex
The proposed algorithm has been evaluated in a simulation environment
to determine its scalability and to showcase the class of difficult
non-monotone problems it can solve. The model of the manipulator used
corresponds to a 7-DOF left arm of a Baxter. The solution paths were
executed on a Baxter robot in open loop trials. The identical objects
are designed as cylinders with height $14cm$ and radius $4cm$. The
class of non-monotone problems (needing multiple grasps of an object),
which the algorithm addresses, have not been solved by any efficient
method to the authors knowledge. Comparisons with the brute-force
exploration has not been performed; a search oninge combined
configuration space of the manipulator and the objects is
computationally inefficient.

\begin{figure}
\centering
\includegraphics[width=0.15\textwidth]{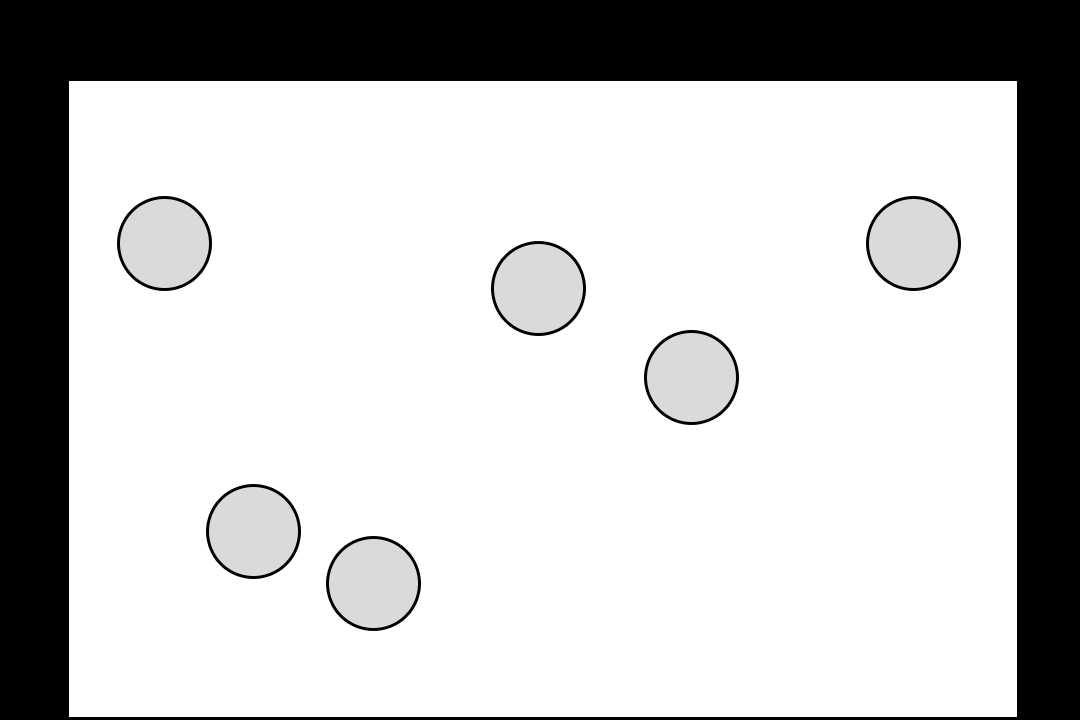}
\includegraphics[width=0.15\textwidth]{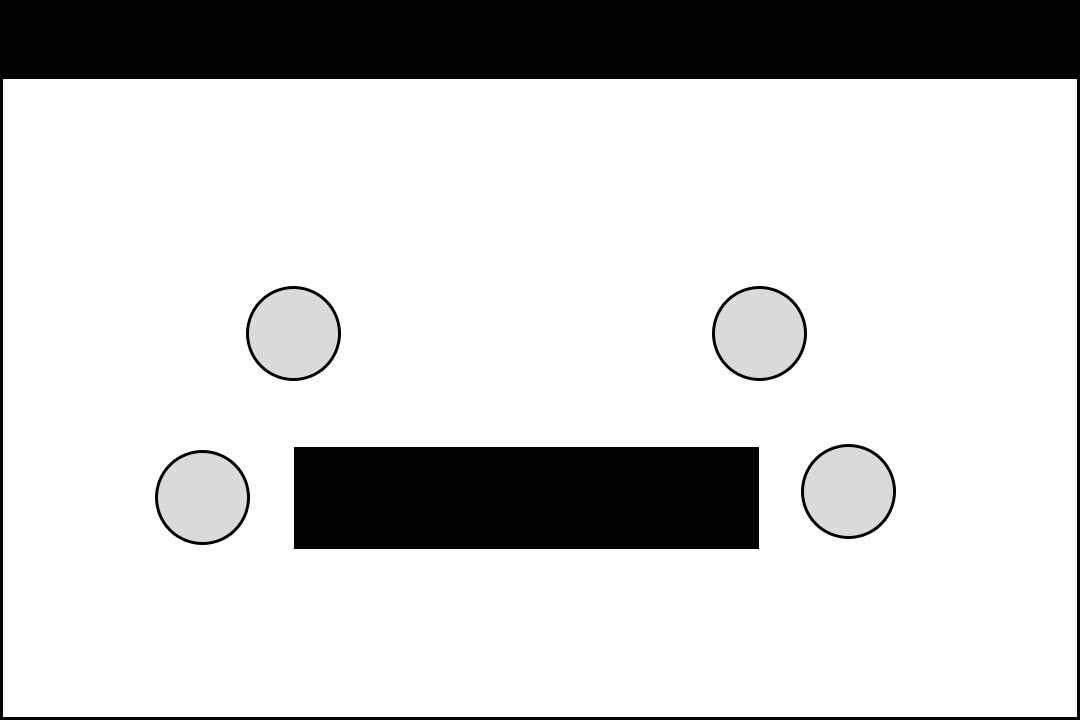}
\includegraphics[width=0.15\textwidth]{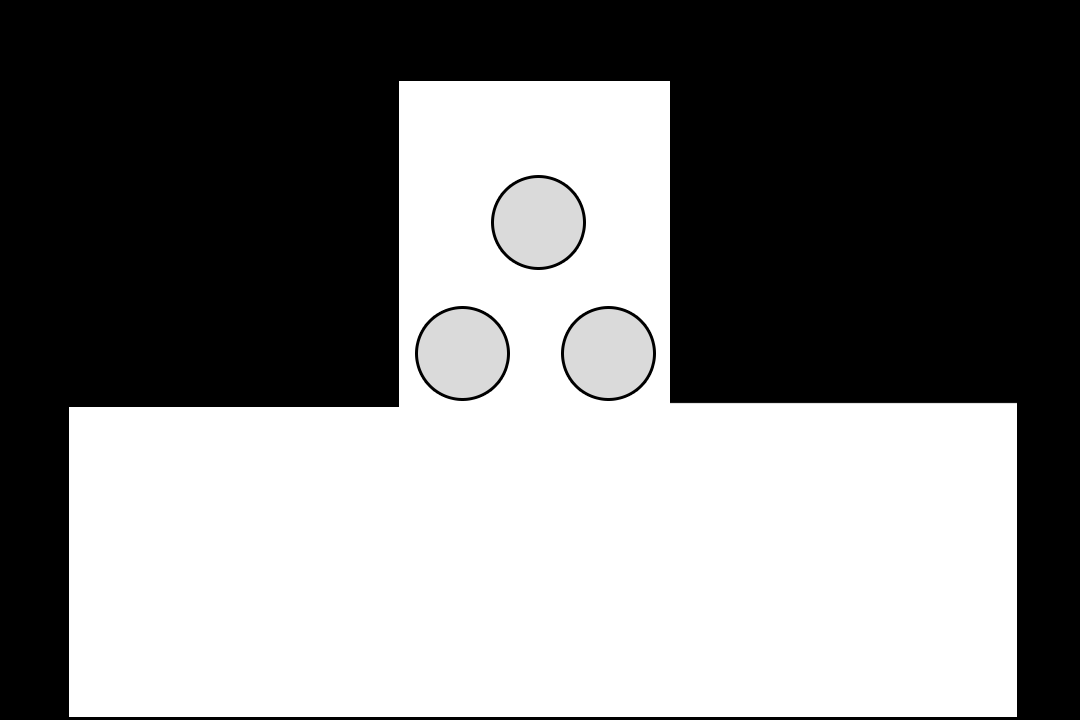}\\
\vspace{0.1in}
\includegraphics[width=0.15\textwidth]{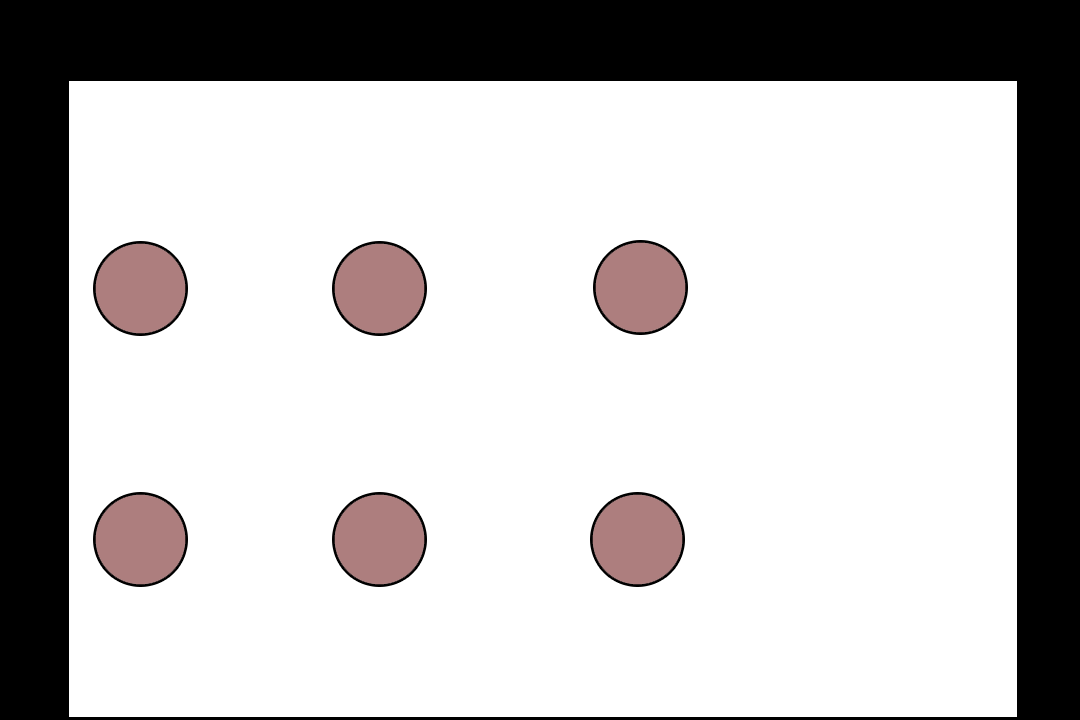}
\includegraphics[width=0.15\textwidth]{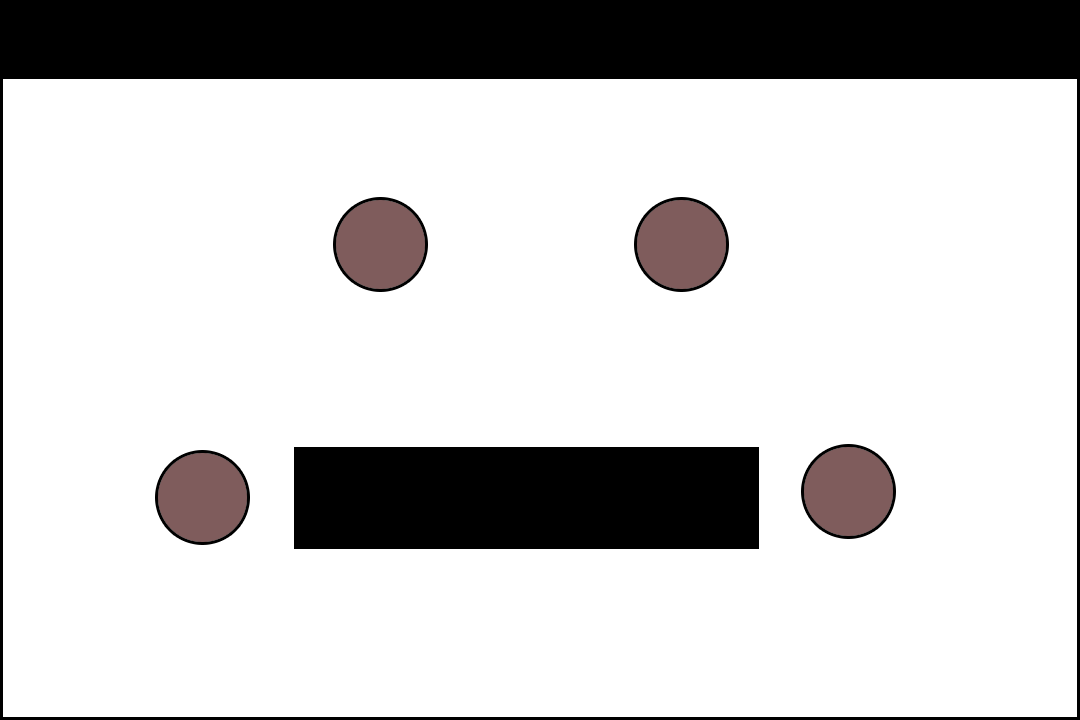}
\includegraphics[width=0.15\textwidth]{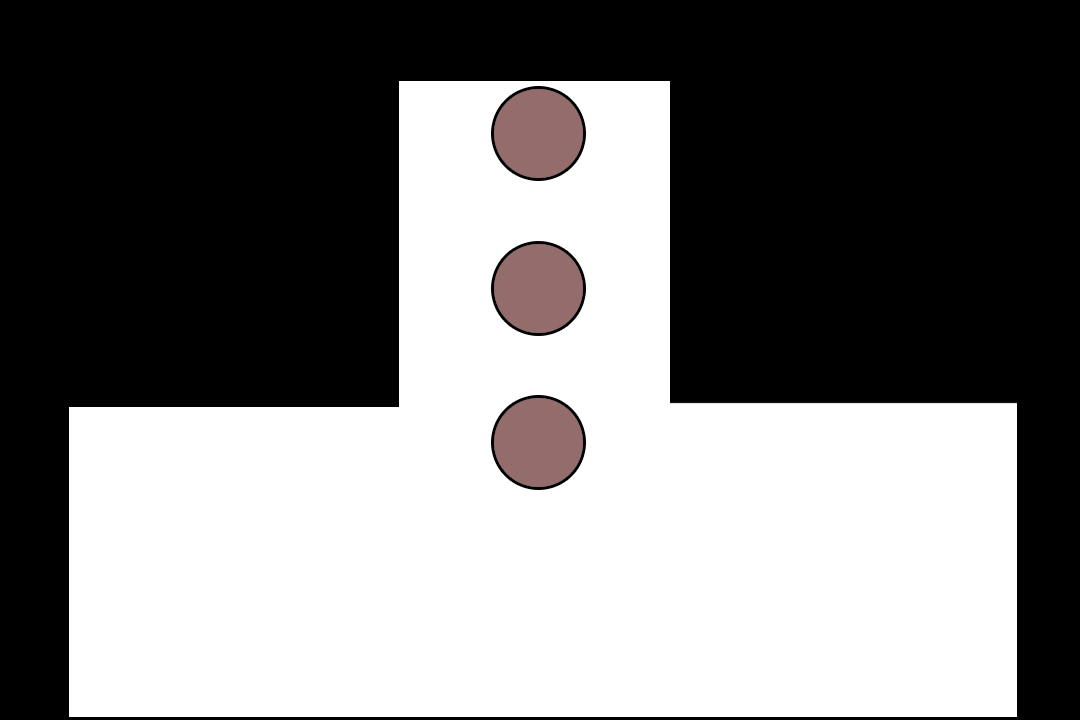}
\caption{The top and bottom rows are the initial and final positions
for the randomized grid problem(left), the non-monotone benchmark
1(middle), the non-monotone benchmark 2(right)}
\label{fig:experimental_setup}
\end{figure}

To evaluate the proposed approach, a series of of rearrangement
problems were considered. These problems include environments with
random initial configurations to a final grid rearrangement of the
objects in a shelf, as well as two non-monotone problems with $\nobj =
3, 4$ (Figure \ref{fig:experimental_setup}). The execution time and
the length of the solution trajectory have been analyzed. These
metrics are reported for different values of $\nblank$, i.e., the
number of ``blank'' poses in the \rpg s. An informed pre-computed
roadmap $\roadmap$ as described in section \ref{sec:implementation}
has been used for the experiments. A trial is a failure if the method
cannot find a solution within the time limit of 600 seconds. The
simulations were performed on a machine running on an Intel(R) Xeon(R)
CPU E5-4650 0 @ 2.70GHz.\\




\noindent \textbf{Randomized grid evaluation: } The first set of
experiments populates a shelf like environment with sets of $2,3,4,6,$
or $8$ objects in mutually exclusive start poses. The goal arrangement
is a uniform grid formation on the shelf. The dimensions of the shelf
prevent the grasping of the objects from the top. This causes objects
to be occluded by other objects in front of them and makes the
rearrangement challenge non-trivial. Different values of $\nblank =
1,2,3,4$ used for these set of experiments. The manipulation roadmap
used for these experiments used $20$ different poses and got $517$
vertices and $6032$ edges.\\


\noindent \textbf{Non-monotone benchmarks: } A second set of experiments 
with 2 non-monotone benchmark problems is used to evaluate the method.
For the first problem 4 objects are placed on a platform which resembles 
the shelf with the sides removed. The goal arrangement requires the two 
front objects to be moved at least twice. The trials are performed for
different values of $\nblank=1,2,3,4$. The same roadmap as before is 
used. 

\begin{figure*}
\centering
\includegraphics[width=0.495\textwidth]{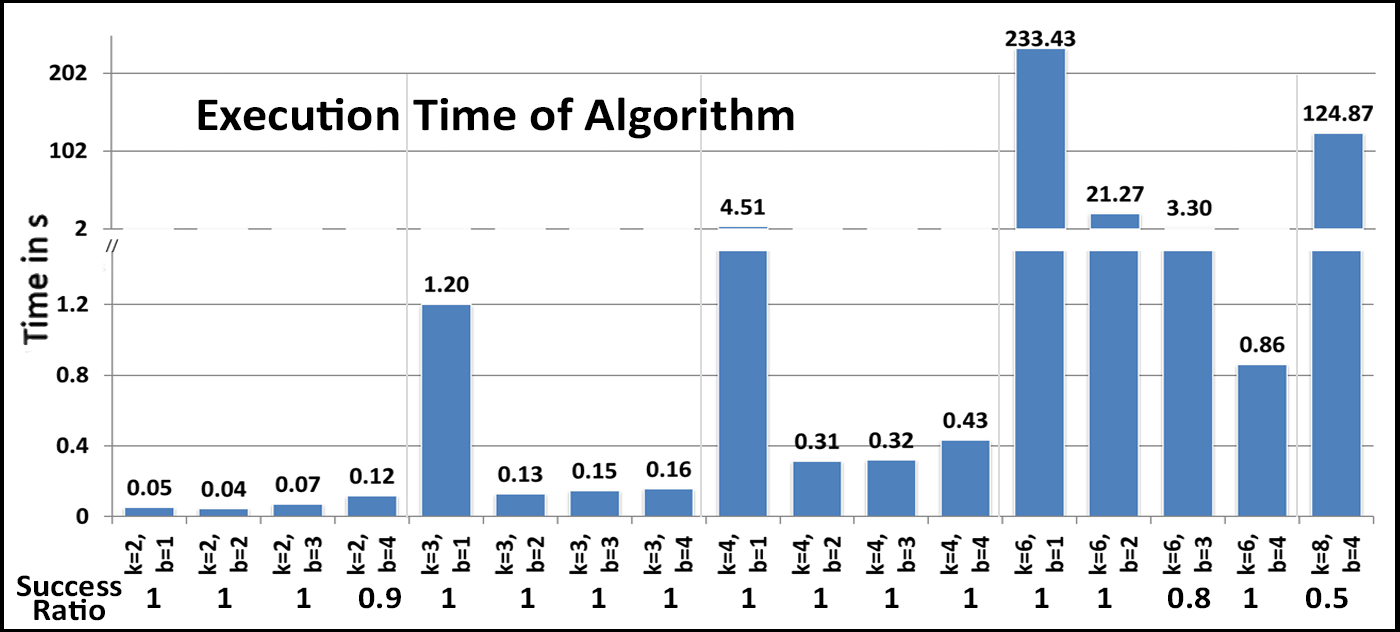}
\includegraphics[width=0.495\textwidth]{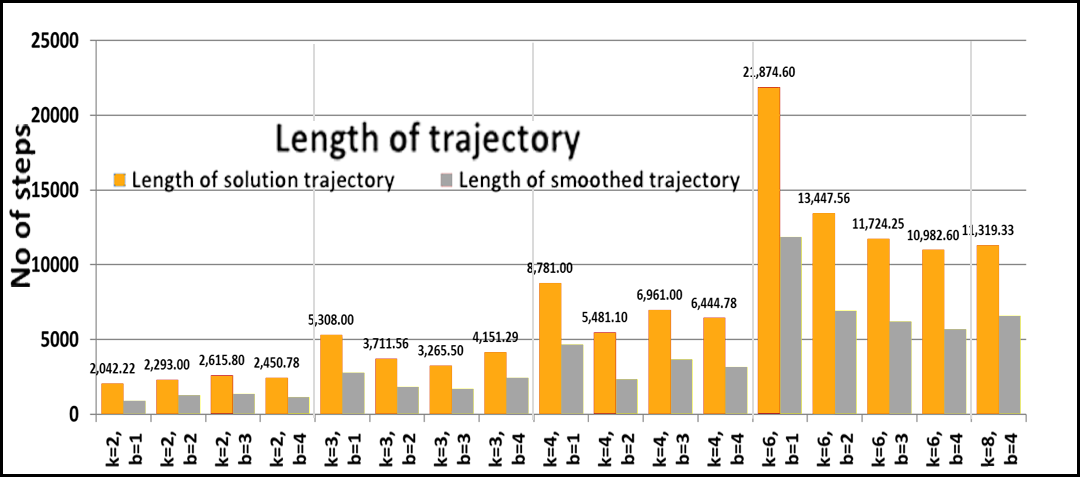}
\caption{(Top) Average running time for the
algorithm to find a solution to the randomized grid problem for
different values of $\nobj$ and $\nblank$. The bottom row indicates
success ratio over 10 trials. (Bottom) Average length of the
trajectory before and after smoothing for the randomized grid
evaluation and different values of $\nobj$ and $\nblank$.}
\label{fig:algo_time}
\end{figure*}

For the second benchmark, 3 objects are placed on a
shelf with two static obstacles that form a narrow cavity between them.
Because of the smaller $\nobj$ only values of $\nblank=1,2,3$ are tested. 
A different roadmap is used because of the constrained nature of the problem. This roadmap
is computed with $30$ poses and constructed with $684$ vertices, $9628$ edges.

\subsection{Execution time} Fig.\ref{fig:algo_time}(top) shows
that the algorithm achieves a high success ratio for
$\nobj=2,3,4,6$. Trials with $8$ objects could not finish within the
stipulated limit of $600s$. $k=8$ causes an increase in the problem
complexity. The case with $8$ objects is constrained by the size of
the shelf in terms of the availability of free poses on the shelf and
free volume required to move. This affects the motion planning
complexity for pose connections. The relatively sparse roadmap used in
the experiment, helps the running time of the problems, but for the
constraining setup, the solution is not always achieved within the
time limit.


The value of $\nblank$ seems to have a significant role in the
execution time for the same $\nobj$. $\nblank=1$ seems to deteriorate
the algorithm's running time for higher values of $\nobj$. This is a
result of a very slow rate of exploration of the poses available on
the shelf for rearrangement, due to only one empty pose available in
the \rpg. However the connectivity in an \rpg\ is maximized. High
values of $\nblank$ introduce a high combinatorial component of
$\nobj$ objects in $\nobj+\nblank$ poses in every \rpg\ and the
connectivity of the \rpg\ decreases. For every value of $\nobj$ the
value of $\nblank$ with the best performance in terms of execution
time, indicates this trade-off.

\subsection{Solution quality} The quality of the solution in the randomized grid experiment, shown in Fig.\ref{fig:algo_time}(bottom). The value of $\nblank=1$ introduces a low exploration rate, which increases the length of the solution. The conservative solution trajectories are shortened by an average of $48\%$.  

\subsection{Connectivity} 
\begin{figure*}
\centering
\includegraphics[width=0.49\textwidth, height=0.3\textwidth]{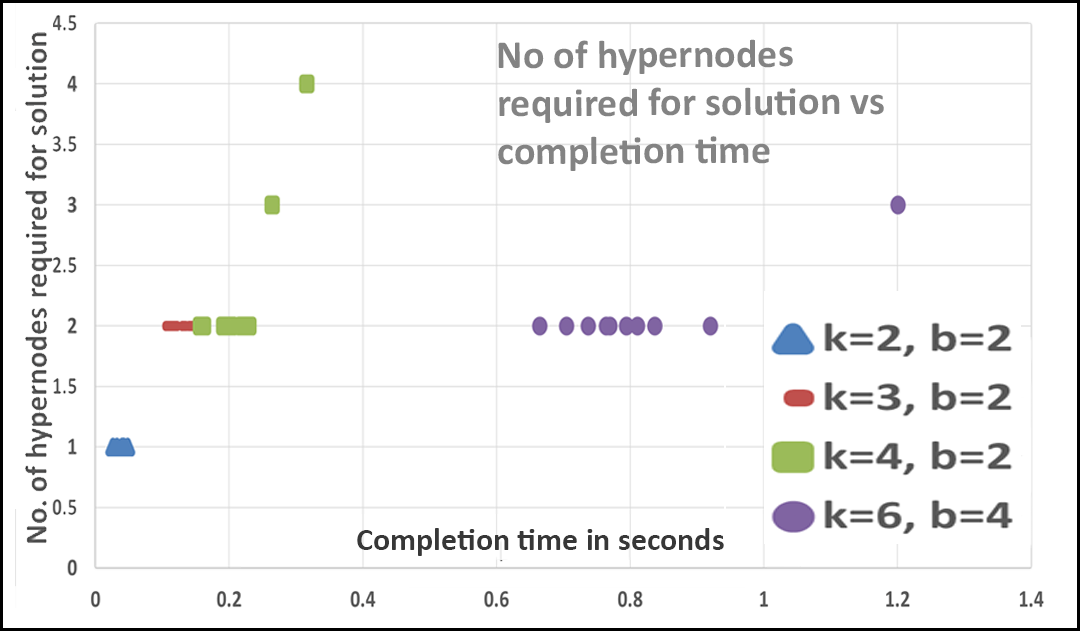}
\includegraphics[width=0.49\textwidth, height=0.3\textwidth]{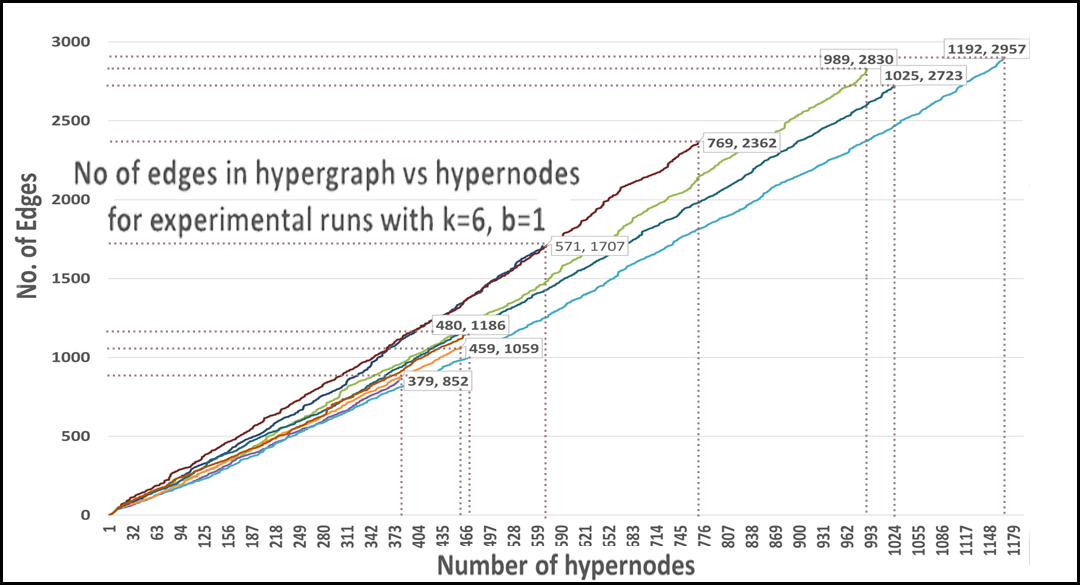}
\caption{The figure on the left shows the amount of time taken and size of the hyper-graph required to achieve the solution, for different runs of the randomized grid problem, for different values of $\nobj$ and $\nblank$. The figure on the right shows the number of edges in the hyper-graph with the creation of hyper-nodes, for $\nobj=6$ and $\nblank=1$.}
\label{fig:nodes_v_time}
\end{figure*}

The key feature of the algorithm is the way it constructs the hyper-graph from the hyper-nodes. The connectivity of this hyper-graph is crucial in finding the solution to the rearrangement problem. Fig.\ref{fig:nodes_v_time}(right) shows the rate at which new edges are created, with the introduction of every new hyper-node, for the different trials for the randomized grid experiment with $\nobj=6$,$\nblank=1$.
The random trials corresponding to the best values of $\nblank$ for
every $\nobj=2,3,4,6$ from Fig.\ref{fig:algo_time} are analyzed for
the number of hyper-nodes that were required for finding the solution
and the time taken. Complex solutions are efficiently discovered with
short execution times. For $\nobj=6$,$\nblank=4$, the solution grasps
every object $2.67$ on average times over all the trials.


\subsection{Benchmark evaluation}

\begin{figure*}
\centering
\begin{minipage}{0.18\textwidth}
\includegraphics[width=0.99\textwidth]{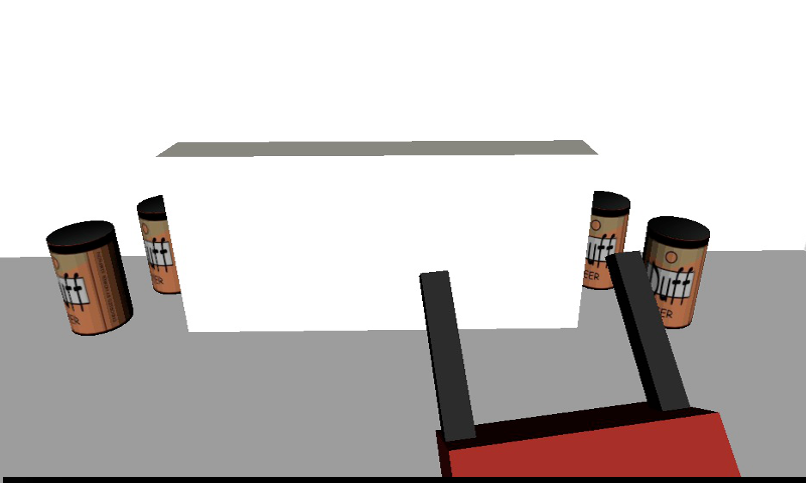}
\includegraphics[width=0.99\textwidth]{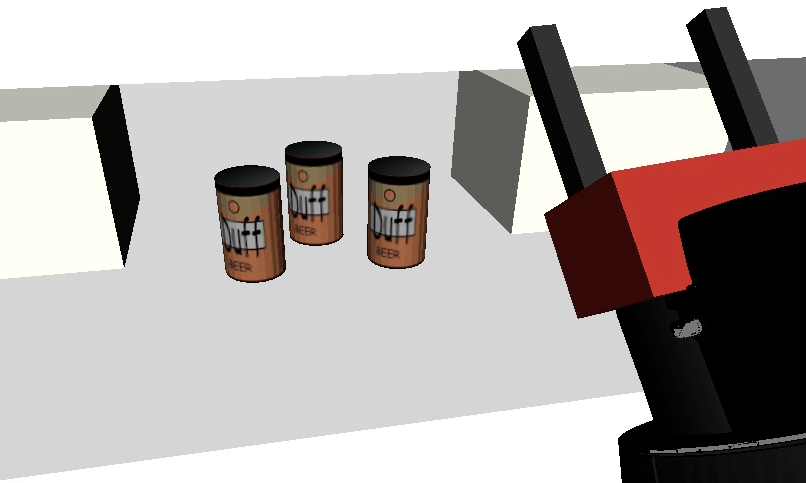}
\end{minipage}
\begin{minipage}{0.80\textwidth}
\includegraphics[width=0.49\textwidth]{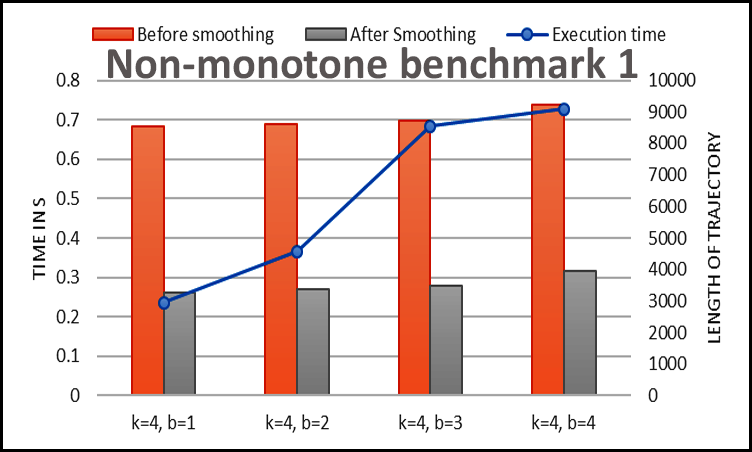}
\includegraphics[width=0.49\textwidth]{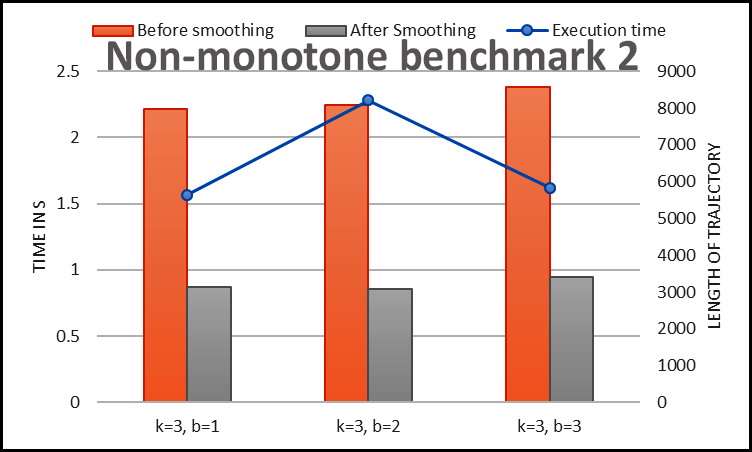}
\end{minipage}

\caption{The simulation environment on the left are for benchmark 1(top) and the benchmark 2(bottom). The performance of the algorithm for the non-monotone benchmark 1(middle) and non-monotone benchmark 2(right) are also shown.}
\label{fig:non_monotone_setup}
\end{figure*}

Fig.\ref{fig:experimental_setup} shows that the objects need to be
grasped at least twice in order to achieve the rearrangement.  The
algorithm is executed $10$ times for different values of
$\nblank$. The algorithm achieves a solution in every trial of the
benchmark problems. The time of execution for both benchmarks
indicates an efficient solution in Fig.\ref{fig:non_monotone_setup}.
The second benchmark is solved on a roadmap with $684$ vertices,
$9628$ edges and $30$ poses. The bigger roadmap, used in this
benchmark, is necessary to deal with the constraining nature of the
problem.  Increasing the size of the roadmap and the number of poses
greatly helps in making more constrained problems solvable.

%

%

%% file: 07_discussion.tex
The proposed method solves rearrangement problems for similar objects,
using a high DOF robot arm, including non-monotone instances that are
hard to address with existing efficient methods \cite{Stilman:2007kl}.
Probabilistic completeness can be argued for a general class of
problems with the proposed method and precomputation can be
appropriately utilized to achieve fast solution times.


The approach can fail in cases where the arm must place an object to a
configuration from where the arm cannot be retracted to a safe
configuration.  Moreover, the algorithm does not address objects with
different labels or geometries. The motivating work on pebble
graphs \cite{Solovey2012k-Color-Multi-R}, however, is providing a
framework to extend the current approach to the case of different
objects. Furthermore, the current implementation takes advantage of
grasp symmetries about the Z axis for cylindrical objects placed
upright. Dealing with general grasps and general resting poses for the
objects needs further investigation. There are also interesting
variations that can be explored in relation to different ways for
connecting hyper-nodes that may not require the two nodes to share $k$
poses.

The paths computed in simulation have been tested in open-loop (for
known initial and target object arrangement) on a real Baxter system
arranging cylindrical objects in a shelf.  Long paths that involve
multiple grasps were unsuccessful as the robot's path deviated due to
errors from the computed one. Nevertheless, small rearrangement
challenges were solved using the real system
\footnote{Videos can be
found: \url{http://www.cs.rutgers.edu/~kb572/wafr_14_results/rpg/}}. 
Future efforts will focus on the computation of robust rearrangement
trajectories and their robust execution given appropriate sensing
input \cite{Levinh:2012kx}.